%% file: main.tex
\documentclass[letterpaper, 10 pt, journal, twoside]{IEEEtran}
\ifCLASSINFOpdf
  \usepackage[pdftex]{graphicx}
\else
\fi
\usepackage{amsmath,amsfonts}
\usepackage{siunitx}
\usepackage{multirow}
\usepackage{booktabs}
\usepackage{csquotes}
\usepackage{xcolor}
\usepackage[table]{xcolor}
\usepackage{hyperref}
\usepackage[capitalize]{cleveref}
\usepackage{threeparttable}
\usepackage{arydshln}  

\begin{document}

%
\title{OMCL: Open-vocabulary Monte Carlo Localization}
%
%
%

\author{Evgenii Kruzhkov$^{1}$, Raphael Memmesheimer$^{1}$, and Sven Behnke$^{1}$%
\thanks{$^{1}$All authors are with the Autonomous Intelligent Systems group, Computer Science Institute VI – Intelligent Systems and Robotics – and the Center for Robotics and the Lamarr Institute for Machine Learning and Artificial Intelligence, University of Bonn, Germany;
        {\tt\footnotesize ekruzhkov@ais.uni-bonn.de}}%
\thanks{© 2025 IEEE. Personal use of this material is permitted. Permission from IEEE must be obtained for all other uses, in any current or future media, including reprinting/republishing this material for advertising or promotional purposes, creating new collective works, for resale or redistribution to servers or lists, or reuse of any copyrighted component of this work in other works}
}
\maketitle

\begin{abstract}
Robust robot localization is an important prerequisite for navigation, but it becomes challenging when the map and robot measurements are obtained from different sensors.
Prior methods are often tailored to specific environments, relying on closed-set semantics or fine-tuned features. 
In this work, we extend Monte Carlo Localization with vision-language features, allowing OMCL to robustly compute the likelihood of visual observations given a camera pose and a 3D map created from posed RGB-D images or aligned point clouds
These open-vocabulary features enable us to associate observations and map elements from different modalities,  and to natively initialize global localization through natural language descriptions of nearby objects.
We evaluate our approach using Matterport3D and Replica for indoor scenes and demonstrate generalization on SemanticKITTI for outdoor scenes. 
The code is accessible at \url{https://github.com/AIS-Bonn/omcl}.
\end{abstract}


%
\IEEEpeerreviewmaketitle

\section{Introduction}
%
%
%
%
\input{chapters/1_introduction}
%
\section{Related Works}\label{reelated-works:all}
\input{chapters/2_related_works}
\section{Method}\label{method:method}
\input{chapters/3_method}

\section{Experiments}
\input{chapters/4_experiments}

\section{Conclusion}
\input{chapters/5_conclusion}
\section*{Acknowledgment}
This work has been partially funded by the Federal Ministry of Research, Technology and Space of Germany (BMFTR) under grant no. 01IS22094A WestAI and within the Robotics Institute Germany, grant no. 16ME0999.

\ifCLASSOPTIONcaptionsoff
  \newpage
\fi



%



\bibliographystyle{IEEEtran}
\bibliography{OMCL.bib}

%








\end{document}

%% file: chapters/1_introduction.tex
\IEEEPARstart{L}{ocalization} is a fundamental problem in robotics, allowing robots to estimate their position and orientation within an environment. 
Traditional approaches utilize single or fused sensor modalities~\cite{krombach2018feature,droeschel2018efficient,cao2024slcf}.
Cross-modal localization~\cite{cattaneo2025cmrnext} further increases flexibility for heterogeneous systems and facilitates map reuse across platforms, while incorporating semantic awareness enhances robot autonomy~\cite{peng2023openscene,huang2023visual}.
The semantics can be incorporated through object detection~\cite{zimmerman2023constructing}, semantic segmentation~\cite{YuLLXY0Q_IROS2018}, or vision-language features~\cite{muhammad2023clip, RadfordKHRGASAM_ICML2021}.
This letter introduces the Open-vocabulary Monte Carlo Localization (OMCL) framework, which extends Monte Carlo Localization (MCL) with vision-language features~\cite{RadfordKHRGASAM_ICML2021}. These abstract features enable affordable camera-only localization in 3D maps created from different sensors, like RGB-D cameras or LiDAR.
\cref{fig:logo} demonstrates the language map and localization process of the proposed framework. 

Our approach stores features learned by contrastive language-image pretraining (CLIP~\cite{RadfordKHRGASAM_ICML2021}) in a 3D map.  The positions of features are represented within a spatial language map, while RGB input is processed to extract the features that describe the current observation.
Ray tracing is employed to estimate the correlation between the observations and the map allowing for pose estimation in 3D environments.
We propose a technique to convert pre-existing maps to the compatible representation for our localization framework, compare it to state-of-the-art methods, and conduct a comprehensive ablation study.
Our contributions include the following:

\input{images/logo}

\begin{itemize}
    \item \textbf{Language-grounded localization.} We present OMCL, a novel localization framework that grounds pose estimation in language features and accelerates global localization via open-vocabulary prompts.
    \item \textbf{Cross-modal sensor usage.} OMCL includes a mapping module that constructs a unified, sparse language map, enabling sensors of different modalities to be used for mapping and localization (e.g., RGB-D or point clouds for mapping and RGB for localization).
    \item \textbf{Generalization.} OMCL is compatible with independently constructed point clouds, enabling reuse of existing maps, and it generalizes across indoor and outdoor environments.
\end{itemize}
OMCL offers an effective, language-feature guided solution for localization. It achieves state-of-the-art performance compared to existing baselines and opens up new possibilities for natural language-guided and cross-modal localization.
A robot can be localized by non-expert users without the necessity of comprehending the underlying map representation through the use of natural language descriptions in OMCL.

%% file: images/logo.tex
\begin{figure}
  \centering
  \includegraphics[width=.7\linewidth]{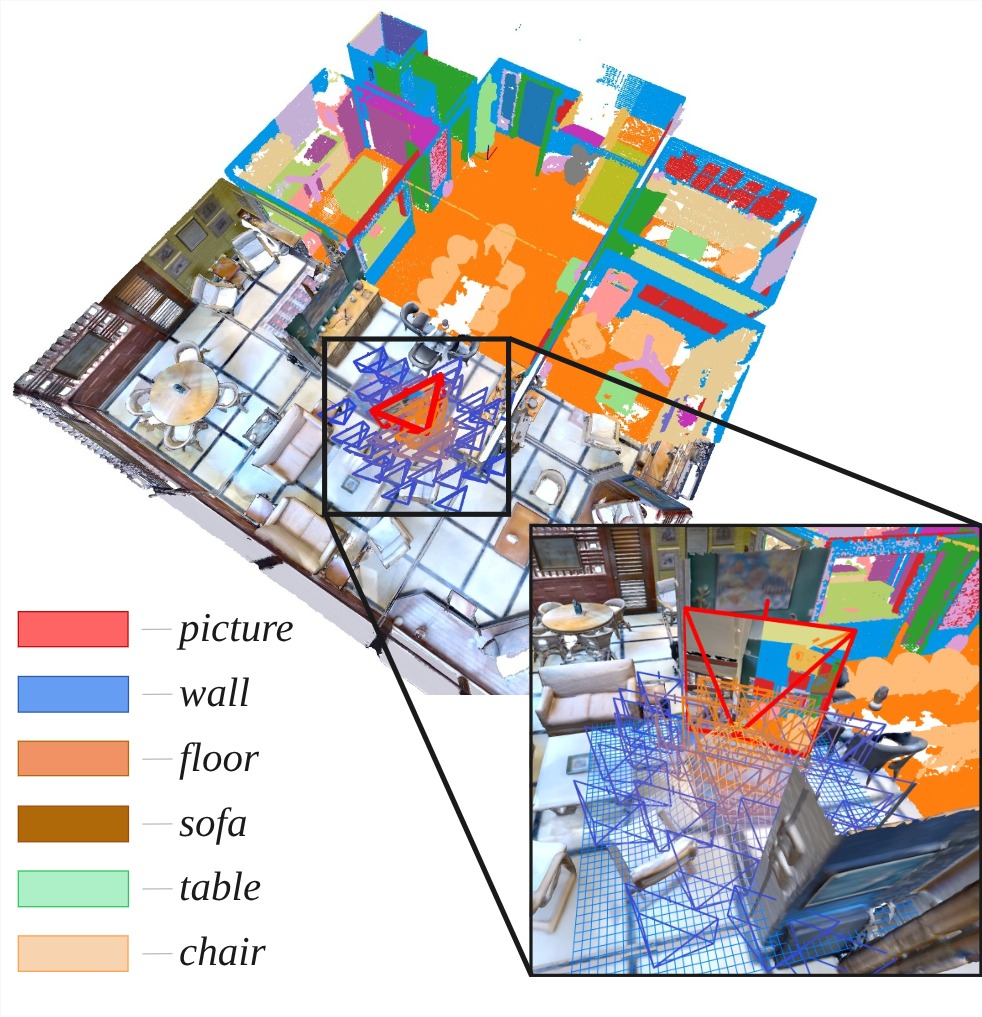}
   \caption{OMCL particles are sampled on a Language Map storing open-vocabulary features.
Each particle represents a candidate camera pose.
Particles are weighted according to how well the VLM-processed RGB input matches the ray-traced map value at the relevant location.
The red particle denotes the estimated pose (weighted mean).
The Language Map is colored by similarity to the prompted labels.
}
\vspace{-0.45cm}
   \label{fig:logo}
\end{figure}

%% file: chapters/2_related_works.tex
\textbf{Semantic Localization.}
We focus on visual continuous localization within 3D maps constructed from cross-modal sensors~\cite{yudin2023cloudvision, zhang2024visual}.
Unlike Loc-NeRF~\cite{maggio2023loc}, which relies on rendered visual consistency for pose tracking, we achieve sensor cross-modality through semantic consistency.
Prior methods, SeDAR~\cite{mendez2018sedar} and SIMP~\cite{zimmerman2023constructing}, use MCL for long-term indoor semantic localization on floor plans; in contrast, we propose a more general map representation and a semantic-consistency measurement model.
We evaluate against SeDAR~\cite{mendez2018sedar} and SIMP~\cite{zimmerman2023constructing} by adapting our map representation to their floor-plan views.

SemLoc~\cite{liang2022semloc} and Zhang \textit{et al.}~\cite{zhang2023cross} explore semantic consistency for localization in outdoor environments.
Unlike these methods, we rely purely on semantic consistency in a multi-modal feature space, without additional geometry-based optimization terms or custom semantic segmentation models.
CMRNext~\cite{cattaneo2025cmrnext} is a state-of-the-art camera-to-LiDAR map-matching baseline for outdoor localization, which, in contrast to our approach, relies on a fine-tuned neural network model.

Unlike our approach, place recognition (PR) methods~\cite{zhao2024pnerfloc} estimate the pose from a single image, whereas we handle continuous image streams and pose tracking, potentially requiring multiple observations before convergence.
However, MCL allows using the same map for both global localization and subsequent pose tracking~\cite{atanasov2016localization}.
While PR can be used for initialization, it requires additional PR-specific infrastructure; in contrast, we propose a map-native, prompt-augmented initialization.

\textbf{Semantic Mapping.}
Many recent works focus on open-set semantic mapping of indoor environments followed by 3D semantic segmentation.
SAM3D~\cite{yang2023sam3d} creates 3D scene masks from SAM~\cite{kirillov2023segment}.
OpenScene~\cite{peng2023openscene} proposes a method to directly predict point-wise CLIP~\cite{RadfordKHRGASAM_ICML2021} features for input point clouds.
ConceptGraphs~\cite{gu2024conceptgraphs} constructs a graph-based representation with incorporated semantic feature vectors.
ConceptFusion~\cite{jatavallabhula2023conceptfusion}, implemented on top of $\nabla$SLAM~\cite{jatavallabhula2020slam}, produces unordered multi-modal maps.
LiLMaps~\cite{kruzhkov2025lilmaps} investigates sequential visual–language mapping for implicit representations, reducing the memory footprint of such dense maps.
Unlike these mapping-focused works, we provide options for constructing the map from different sensing modalities and couple it with a ray-tracing-based, semantic-aware probabilistic localization module.

The recent OVO~\cite{martins2025open} mapping approach uses Gaussian-SLAM~\cite{yugay2312gaussian} and ORB-SLAM2~\cite{mur2017orb} but does not exploit the semantic information added to the map to improve pose estimation.
Concurrent to our work, RayFronts~\cite{alama2025rayfronts} similarly constructs an ordered map by averaging visual–language features from multiple directions. However, they focus on multi-modal open-set querying and beyond-range semantic classification.
Similar to the aforementioned works, we construct our map representation from externally estimated poses, while tight integration with simultaneous localization and mapping (SLAM)~\cite{mccormac2018fusion++, stuckler2012semantic} is a possible future improvement.

\textbf{Vision Language Models.}
We employ contrastive Vision-Language Models (VLMs)~\cite{RadfordKHRGASAM_ICML2021,peng2023openscene,li2022language,alama2025rayfronts,zou2023generalized}, for the localization task.
VLMs and large language models (LLMs) are widely used for related navigation~\cite{chen2023open,huang2023visual} and scene understanding~\cite{ha2022semantic} tasks.
We use pixel-wise visual-language features (such as CLIP~\cite{RadfordKHRGASAM_ICML2021}) produced by the image encoder as open semantic representations in the maps.
By default, our framework employs LSeg~\cite{li2022language} and OpenScene~\cite{peng2023openscene} encoders, but we demonstrate flexibility with respect to the choice of visual–language backbone by switching to X-Decoder~\cite{zou2023generalized} and NARADIO~\cite{alama2025rayfronts}.

We don't compete with the aforementioned approaches; instead, our method creates a unified visual-language map from cross-modal sensors and evaluates its downstream impact on localization.

%% file: chapters/3_method.tex
\input{images/pipeline}

The proposed pipeline is illustrated in~\cref{fig:pipeline}.
The map representation constructed and used by our framework is the Octree Language Map.
We focus on storing visual–language features~\cite{RadfordKHRGASAM_ICML2021,li2022language,peng2023openscene} in the map to investigate their potential for advancing perception systems.
~\cref{method:mapping} covers Octree Language Map implementation details and, after that, describes its construction from different inputs based on pixel-wise visual encoders for RGB-D input and OpenScene~\cite{peng2023openscene} for point clouds.

In~\cref{method:localization} we address the visual-only localization in 3D maps.
Although multimodal sensor configurations (e.g., RGB-D cameras, LiDAR) can be employed for mapping, the subsequent localization requires only visual data and map-scaled odometry (Motion Model), under the assumption that the features in the map are consistent with those extracted from the input RGB image.

We focus on mapped visual–language features that are semantically grounded using open-set semantic prompts.


\subsection{Mapping}\label{method:mapping}
Throughout the mapping process, we generate a volumetric language map (Octree Language Map), a sparse octree-based structure, in which each map voxel is linked to the corresponding $F$-dimensional visual-language feature~\cite{RadfordKHRGASAM_ICML2021,li2022language,peng2023openscene}.
The Octree Language Map offers compact storage at fine-grained resolution and provides efficient ray-tracing functionality~\cite{KaolinLibrary}.
It retains only features that are mutually different by at least a cosine distance threshold $\tau$.
Each voxel in the map stores the index of its corresponding feature in the database \textrm{Features$_{\textrm{DB}}$}:
\begin{equation}
\begin{aligned}
&\textrm{Features$_{\textrm{DB}}$} = \{ f_i \in \mathbb{R}^n,  \mid d(f_i, f_j)>\tau; \; i,j \in \mathbb{N}\}, \\
&d(A, B) = 1-\dfrac{A\cdot B}{\lVert{A}\rVert \lVert{B}\rVert},\ {\text{where}}\ A,B \in \mathbb{R}^n.
\end{aligned}
\label{eq:system_of_equations}
\end{equation}

We provide two mapping approaches: one for posed RGB-D images and another for fused point clouds (\cref{fig:pipeline}).
The produced Octree Language Map is unified.

\textit{Input Option 1.} Unless stated otherwise, we use the LSeg~\cite{li2022language} visual–language model to extract pixel-wise features and project them onto the Octree Language Map using depth measurements and known camera poses.
The mapping is sequential, and features projected into the same voxel are cumulatively averaged with the existing ones.
Once all data have been processed, the mapped features form \textrm{Features$_{\textrm{DB}}$} according to~\cref{eq:system_of_equations}.
The first features that satisfy~\cref{eq:system_of_equations} are added to \textrm{Features$_{\textrm{DB}}$}, while subsequent ones are replaced by their cosine-closest counterparts from \textrm{Features$_{\textrm{DB}}$}.


\textit{Input Option 2.} We process the aggregated point cloud map with the OpenScene 3D distillation model~\cite{peng2023openscene} to predict visual–language features for each point in a feed-forward manner.
The Octree Language Map is then formed by averaging features within voxels and constructing \textrm{Features$_{\textrm{DB}}$} as input option 1.
The described approach is suitable for converting existing 3D maps into our Octree Language Map representation.

\textit{Semantic Grounding.}
Each element of the constructed \textrm{Features$_{\textrm{DB}}$} can be viewed as an automatically created semantic class.
We propose to ground \textrm{Features$_{\textrm{DB}}$} using user-defined open-set semantic prompts.
In practice, we employ the text encoder corresponding to the mapping model (CLIP~\cite{RadfordKHRGASAM_ICML2021} for LSeg~\cite{li2022language} and OpenScene~\cite{peng2023openscene}) and redefine \textrm{Features$_{\textrm{DB}}$} using the features of the provided open-set semantic classes.
The feature indices stored in the voxels of the Octree Language Map are remapped based on the highest cosine similarity between their original associated features and the redefined \textrm{Features$_{\textrm{DB}}$}.
Old features that are too far in cosine distance from the provided open-set semantic classes are discarded along with their corresponding map voxels.
Grounding reduces the memory footprint by decreasing the number of stored features. It increases the discriminability of the remaining features and allows the user to specify which semantic classes are represented in the map for downstream localization.


\subsection{Localization}\label{method:localization}
We use RGB images and odometry (Motion Model) to localize within the language maps employing MCL.
We initially assign uniformly distributed weights to each particle and sample particles with random poses around the probable starting position.
When a new RGB image is received, the motion model is applied to all particles to predict their new poses. 
A coarse odometry from any built-in sensor can be employed.
The image is processed using the language-driven semantic segmentation model~\cite{li2022language} to extract per-pixel features.
To form the measurement model of our MCL and assign new weights for the particles, we evaluate the consistency between the extracted features and the map  (\cref{fig:pipeline} bottom).

\textit{Observation Likelihood.}
If the camera is exactly at a particle’s pose, the features extracted from the image should match those stored in the Octree Language Map where viewing rays hit a surface. 
We employ camera intrinsics and each particle's predicted pose to form the rays in world coordinates.
Ray-tracing~\cite{KaolinLibrary} on the map finds the first voxel hit by each ray, from which we retrieve the stored language feature.
Let $\gamma$ and $\varphi$ denote the voxel features and pixel features corresponding to the same rays, respectively. The weight $w^{t}_{i}$ of particle $i$ at time $t$ is then estimated as:
\begin{equation}
w^{t}_{i} = \frac{w^{t-1}_{i} \max(\mathcal{L}_{i},0)}{\sum_{i}({w^{t-1}_{i} \max(\mathcal{L}_{i}},0))},
\;\;
\mathcal{L}_{i} = \frac{1}{N} \sum_{j=1}^{N} \dfrac{\varphi_j\cdot {\gamma}_j}{\lVert{\varphi_j}\rVert \lVert{{\gamma}_j}\rVert},\label{eq:vl-loss} 
\end{equation}
where $N$ represents the total number of rays per particle.

$\mathcal{L}_{i}$ is, in effect, the averaged cosine similarity between image and map features.
The more similar the observations and the map are, the higher the value of $\mathcal{L}_{i}$ and, consequently, the higher the corresponding particle weight $w^{t}_{i}$.
The final camera pose is then estimated as the average of particles that are close to the most likely particle, using their corresponding weights $w^{t}_{i}$.

\begin{figure}
\vspace{0.1cm}  
  \centering
  \includegraphics[width=1.\linewidth]{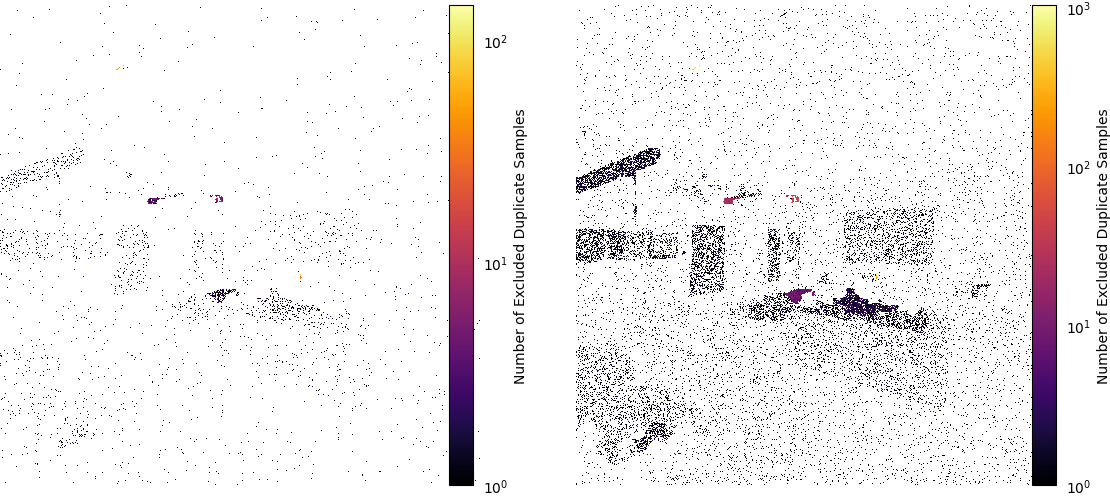}

   \caption{Sampled pixels for images of resolution $540 \times 540$. 
   Both images employ the same sampling masks.
   The left image corresponds to $2^8$ samples per cluster and the right one to $2^{11}$.
   Small clusters have a higher sampling density, but the total number of samples is less for them because the duplicates are discarded.
   }
   \vspace{-0.2cm}
   \label{fig:masks}
\end{figure}

\textit{Stratified Ray Sampling.} 
We reduce computational costs by sampling and using only $N$ pixels and their corresponding features from the image.
However, large surfaces (e.g., walls, floor) occupy most pixels, while smaller but distinctive objects cover far fewer pixels, causing many objects to be underrepresented or missed when sampled uniformly.
Since accurate pose estimation relies on preserving information from all objects, we introduce sampling masks for the image, where each mask corresponds to a cluster of pixels with similar visual-language features.
From each such feature-consistent cluster, we uniformly sample an equal number of pixels, discarding duplicates (\cref{fig:masks}).
The same samples are employed for all particles.
The proposed stratified per-cluster sampling improves pose estimation accuracy, as confirmed by our ablation study in~\cref{experiment:number-of-particles}.

To form the clusters, we use the map features \textrm{Features$_{\textrm{DB}}$} as centroids because, according to~\cref{eq:vl-loss}, localization assumes that map and image features are correlated and \textrm{Features$_{\textrm{DB}}$} already stores only distinct features (see~\cref{eq:system_of_equations}).
The clusters are then formed from the extracted image features assigned to the centroids with which they produce the highest cosine similarity.
Unless stated otherwise, we use full \textrm{Features$_{\textrm{DB}}$} for clustering, though random subsets of centroids may be used to limit the number of clusters and reduce computation.
Both \textrm{Features$_{\textrm{DB}}$} and the clusters are created automatically, requiring no manual labeling for mapping or localization.

\subsection{Prompt-augmented Initialization}\label{prompt-global-loc}

Standard approaches for global localization either require geometric coordinates as input for initialization or sample particles across the entire environment.
However, sampling across the whole environment is inefficient, while geometric coordinates are difficult for non-expert users to interpret or provide, and localization may diverge if the initialization is incorrect.
Moreover, global localization that is decoupled from the subsequent pose tracking step may require storing additional information within the map itself.

\begin{figure}
\vspace{0.1cm}  
  \centering
  \includegraphics[width=.85\linewidth]{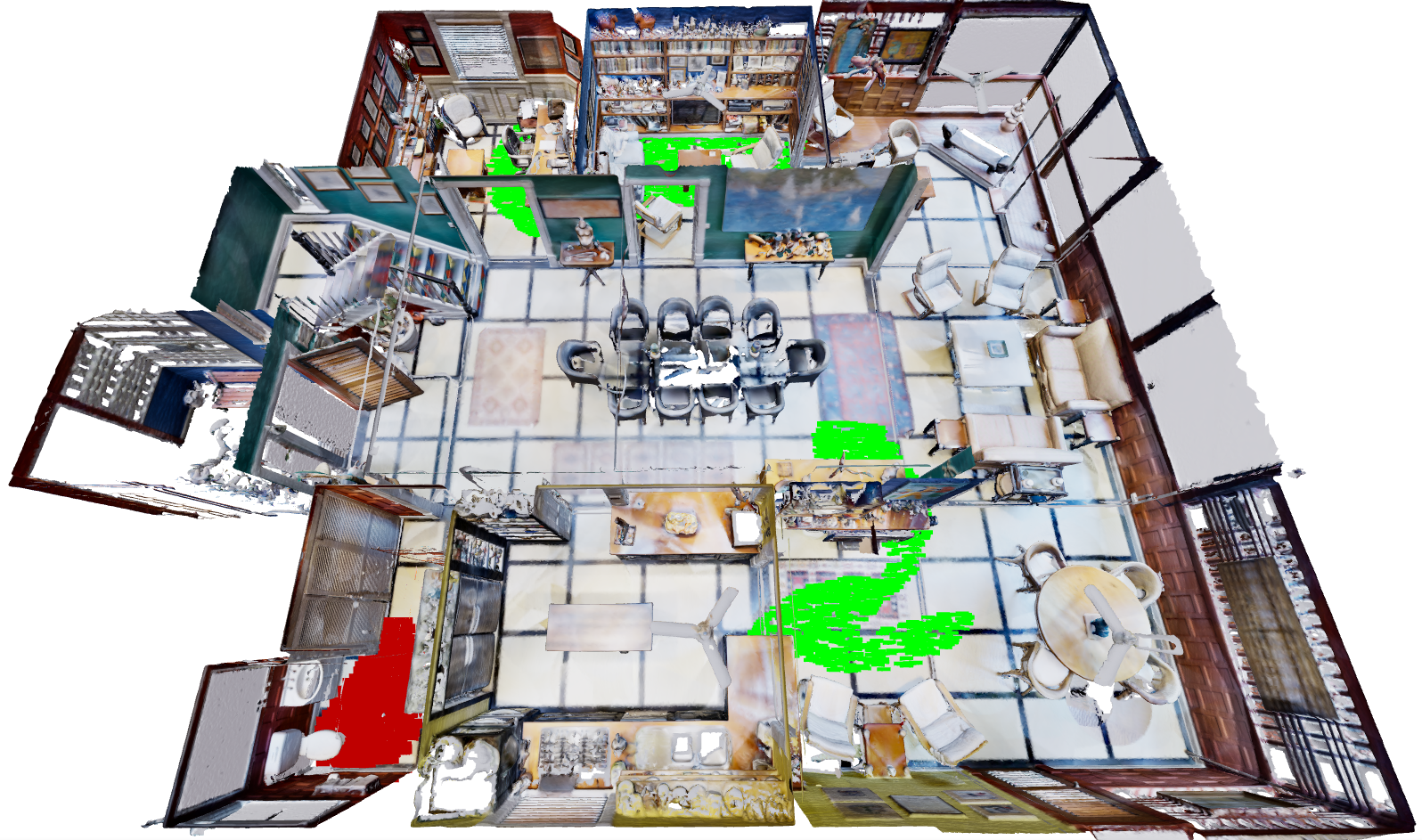}

   \caption{
   An example of initial locations for the global localization based on the user prompt.
   The red spots correspond to the prompt \textbf{\textcolor{red}{(toilet, mirror, towel, sink)}} and the green one to \textbf{\textcolor{green}{(table, chair, picture, door, tv monitor)}}. OMCL particles can be initialized nearby the prompt matching spots instead of the random locations.
   }
   \vspace{-0.5cm}
   \label{fig:init-loc}
\end{figure}

We propose using textual prompts $\theta$  to describe probable initial positions.
As demonstrated in~\cref{fig:init-loc}, the prompt is an open set of natural language words that describe the surrounding environment.
There is no restrictions on word content, as each word is encoded into a feature using a text encoder (CLIP~\cite{RadfordKHRGASAM_ICML2021} for LSeg~\cite{li2022language} and OpenScene~\cite{peng2023openscene}).

First, the proposed initialization accepts natural language words as input, taking a step toward full natural language-based localization.
Second, since the input is text, it is naturally compatible with the output of large language models, enabling initialization with AI agents.
Finally, instead of relying on a single initial location, we initialize particles by sampling them uniformly over all locations that match the given prompt.
Through direct integration with MCL, the initialization subsequently converges to a precise pose, as described in~\cref{method:localization}, based on multiple consecutive observations, and then MCL continues to track the pose.
The same Octree Language Map is employed for both global localization and following pose tracking.

We segment the scene into two classes: floor and surroundings.
We obtain the floor voxels by comparing all stored features with the feature corresponding to the word \enquote{floor}, extracted by the text encoder.  
For each floor voxel, we then find surrounding features $\bar{\gamma}$ within a radius $R$.
Next, for each word in the prompt $\theta$, we count the number of these surrounding features whose cosine similarity with that word exceeds the threshold $\rho$:

\begin{equation}
    \mathcal{V}_{m} = \sum_{n}{(\textrm{CosineSimilarity}(\bar{\gamma}_{n\times F}, \bar{\theta}_{m\times F}) > \rho)},
    \label{eq:variable_v}
\end{equation}
where $n$ is the number of neighbor surroundings, $m$ is the number of words in the prompt, and $F$ is the feature dimensionality.
 
Finally, we estimate the floor voxel–prompt alignment ratio $s$ as a measure of how well the floor voxel matches the prompt, by requiring that at least $k$ of its surrounding voxels correspond to each word:
\begin{equation}
s = \frac{1}{m}\sum_{m}(\mathcal{V}_{m} \geq k), \quad 0 \leq k \leq m, \label{eq:loc-alignment-s} 
\end{equation}
where $s$ is equal to $1$ for floor voxels that match all words.
The particles are uniformly sampled above \enquote{floor} voxels that match all words. 


%% file: images/pipeline.tex
\begin{figure*}[t]
\vspace{0.1cm}  
  \centering
  \includegraphics[width=0.99\linewidth]{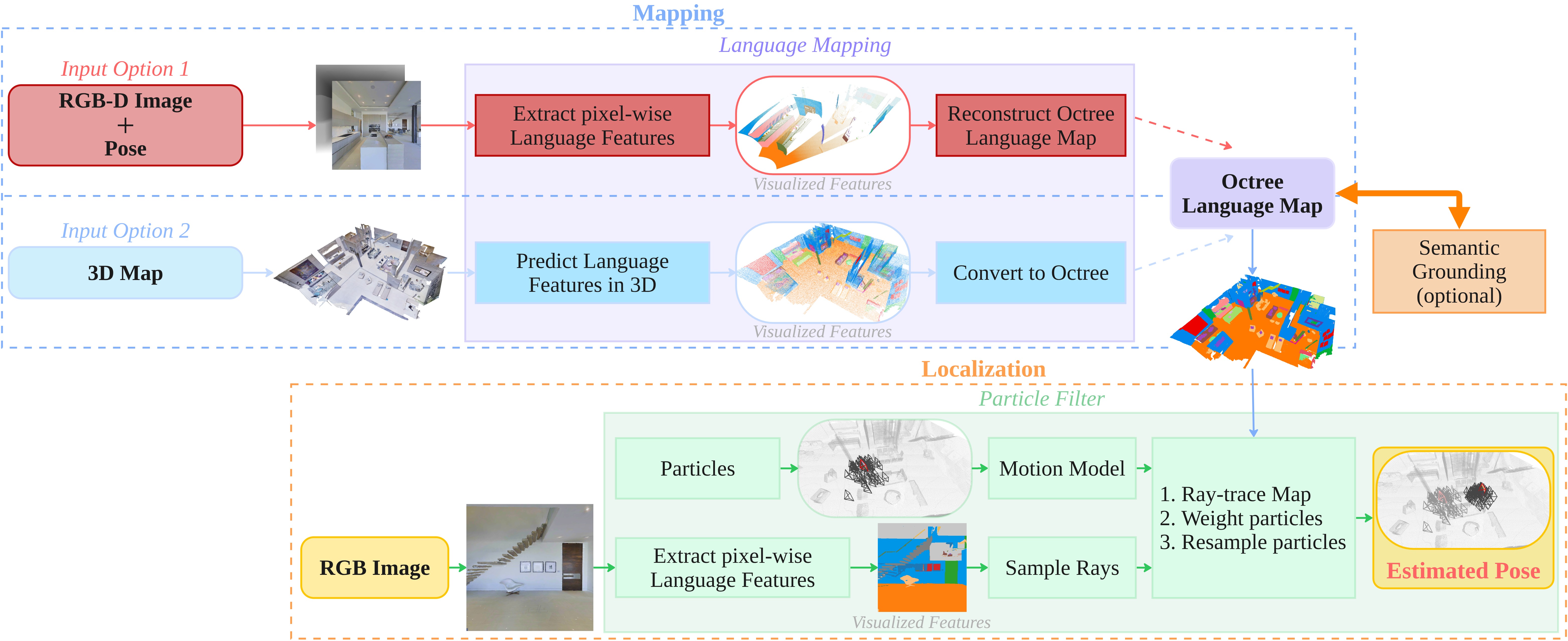}
   \caption{
   \textbf{Mapping:} We propose two options to create Octree Language Maps. 
   \textit{Input Option~1}: OMCL derives language features from RGB images and reconstructs a 3D map from them using the corresponding volumetric data (depth images, LiDAR measurements, etc.).
   \textit{Input Option~2}: Language features are directly predicted on precomputed 3D point clouds for each point and subsequently converted into the octree representation.
   \textbf{Localization:} A particle filter uses an RGB image as the only input, weighting particles by the discrepancy between language features extracted from the input image and those ray-traced from the Octree Language Map.
   Our stratified ray sampling strategy compensates for the imbalance between different object instance sizes in the image. All features are colored for visualization purpose only.
   }
   \label{fig:pipeline}
   \vspace{-0.3cm}
\end{figure*}

%% file: chapters/4_experiments.tex
\input{tables/localization2}
We provide complete evaluation of our localization in medium-size indoor environments on Matterport3D dataset~\cite{Matterport3D}, comparing against similar approaches and using publicly available sequences provided by VLMaps~\cite{huang2023visual}.
We further analyze how localization performance depends on measurement–map consistency.

Using the Replica dataset~\cite{straub2019replica} for smaller, room-sized environments, we obtain localization accuracy aligned with that observed on medium-sized environments and additionally assess the semantic coverage of our Octree Language Map relative to open-vocabulary baselines.
For the large-scale evaluation, we employ the SemanticKITTI urban driving dataset \cite{BehleyGMQBGS:IJRR21}, demonstrating generalization capabilities of the proposed approach.


Unless otherwise noted, we use the parameters summarized in~\cref{tab:parameters}.
We use the lower index notation OMCL$_{\textrm{Mapping}\backslash\textrm{Localization}}$ to indicate the specific model used for mapping and localization, respectively. 
For instance, OMCL$_\textrm{GT}$ means that ground truth semantic images are used for both mapping and localization; while OMCL$_{\textrm{Open}\backslash\textrm{LSeg}}$ means OpenScene~\cite{peng2023openscene} is used for mapping and LSeg~\cite{li2022language} for localization.
OMCL always operates in the continuous feature-embedding space, processing GT labels with a corresponding text encoder.


By default, we ground $\textrm{Features}_\textrm{DB}$ as described in~\cref{method:mapping} with a prompt consisting of the datasets’ semantic class labels.
For reference purposes, we provide evaluation for the ungrounded OMCL$^\textrm{auto}$ version.
All results are obtained using an Nvidia RTX 3090 GPU equipped with 24GB of memory.


\input{tables/parameters}

\subsection{Long-Term Indoor Localization}
\textbf{Localization Evaluation:~\cref{tab:localization}.} We compare our approach with SeDAR~\cite{mendez2018sedar} and SIMP~\cite{zimmerman2023constructing} which, similar to us, employ MCL and measure semantic consistency for long-term indoor localization (\cref{reelated-works:all}).
For fair comparison, all approaches are evaluated on the same Octree Language Map which was created from ground truth data (GT) and projected to a floor plan map (2D) by averaging the data along the vertical dimension.
We denote baselines adapted to the common map as SeDAR$^{\ast}$ and SIMP$^{\ast}$.
The same noisy odometry is applied to all methods: ground-truth poses are perturbed with zero-mean Gaussian noise with standard deviations of 0.10\,m (translation) and \SI{6}{\degree} (rotation). For reference, the average per-step ground truth motion is 0.13\,m in translation and \SI{8}{\degree} in rotation.
Particles are initialized around the starting pose with standard deviations of 0.3\,m and \SI{17}{\degree}.
RTAB-Map~\cite{labbe2019rtab} is employed as the evaluation baseline of our full pose estimation on the Octree Language Map in 3D.
We use the RGB-D variant of RTAB-Map to avoid the degeneracies of monocular SLAM in low-parallax, in-place rotation segments widely present in the dataset.

\input{tables/mapping}
\input{images/APE_trajectories}


For completeness, we report the full APE statistics~\cite{grupp2017evo} in~\cref{tab:localization}, averaged over all sequences, and provide the corresponding measurement–map consistency evaluation in~\cref{tab:mapping}.
We evaluate both OMCL configurations shown in (\cref{fig:pipeline}).
Under \enquote{Input Option 1}, we use OMCL$_\textrm{GT}$ and OMCL$_\textrm{LSeg}$.
Under \enquote{Input Option 2}, OMCL$_{\textrm{Open}\backslash\textrm{LSeg}}$ employs the OpenScene~\cite{peng2023openscene} 3D distillation model for the mapping and, because OpenScene does not accept image input, we use LSeg for localization.
This is feasible because both models operate in the same CLIP feature space.

\textbf{Measurements-Map Consistency Evaluation:~\cref{tab:mapping}.}
We report accuracy, precision, recall and IoU between the map and the observed features during localization, averaged over all poses and sequences for different datasets.
To compute these metrics, we treat each feature stored in the Octree Language Map (i.e., each element of $\textrm{Features}_{\textrm{DB}}$) as an independent semantic class.
For each per-pixel image feature, we identify its corresponding map feature via ray casting and count a correct correspondence if that map feature attains the highest cosine similarity among all features in $\textrm{Features}_{\textrm{DB}}$.

The evaluation quantifies a lower bound on consistency between the map and measurements during localization and assesses both the map’s ability to preserve information and the image encoder’s consistency across viewing directions.
It is a lower-bound because correspondences are discretized as correct/incorrect, whereas localization operates in a continuous feature-similarity space.
Note that localization depends solely on this consistency and not on the absolute correctness of the map’s semantics.

\textbf{Results and Comparison:~\cref{tab:localization,tab:mapping}.}
The metrics reported in~\cref{tab:mapping} align with the localization results in \cref{tab:localization}.
The metrics of OMCL$_\textrm{LSeg}$ are close to those of OMCL$_\textrm{GT}$ and both outperform SeDAR$^{\ast}_\textrm{GT}$ and SIMP$^{\ast}_\textrm{GT}$.
OMCL$_{\textrm{Open}\backslash\textrm{LSeg}}$ lacks sufficient mapping consistency to complete all test sequences; nevertheless, it still outperforms the baselines and successfully executes most sequences.
Moreover, OMCL$_{\textrm{Open}\backslash\textrm{LSeg}}$ can create the Octree Language Map directly from existing point clouds.
The ungrounded OMCL$^\textrm{auto}_\textrm{LSeg}$ achieves metrics comparable to OMCL$_\textrm{LSeg}$.

Larger gaps between the Mean and RMSE in~\cref{tab:localization} indicate more significant deviations from the real pose.
The geometry-based RTAB-Map~\cite{labbe2019rtab} loses tracking in narrow, feature-poor areas.
Although it subsequently relocalizes, the accumulated errors prevent accurate recovery of the full trajectory.
\cref{fig:APE_trajectories} illustrates the performance of OMCL in a variety of scenarios, including loopy paths, corridors, and monotonic trajectories.
\begin{table}[t]
\centering
\begin{threeparttable}
\caption{Replica 3D Semantic and Localization Evaluation.}
\label{tab:replica_metrics}
\renewcommand{\arraystretch}{1.1}
\begin{tabular}{lccccc}
\toprule
 & \multicolumn{2}{c}{All} & \multicolumn{2}{c}{Tail} & \multirow{2}{6em}{\centering Camera pose /\\ATE RMSE [cm]}\\
\cmidrule(lr){2-3} \cmidrule(lr){4-5}
 & \textbf{mIoU} & \textbf{mAcc} & \textbf{mIoU} & \textbf{mAcc} & \\ 
\midrule
OpenScene~\cite{peng2023openscene} & 15.9 & 24.6 & 1.5 & 6.7 & N/A \\
ConceptGraphs~\cite{gu2024conceptgraphs} & 16.7 & 33.7 & 4.4 & 26.8 & N/A \\ 
\underline{RayFronts}~\cite{alama2025rayfronts} & \underline{27.7} & \underline{54.5} & \underline{17.6} & \underline{41.3} & N/A \\
OVO~\cite{martins2025open} & 27.1 & 39.1 & 12.1 & 19.6 & \textbf{0.6 -- 1.9} \\
\textbf{OMCL$_{\textrm{NAR}}$ (ours)} & \textbf{32.1} & \textbf{56.2} & \textbf{21.4} & \textbf{43.9} & \underline{35} \\
\bottomrule
\end{tabular}

\end{threeparttable}
\vspace{-0.3cm}
\end{table}

\subsection{Datasets Generalization}
\textbf{Replica.} The Replica dataset~\cite{straub2019replica} contains photo-realistic 3D indoor scene reconstructions at room and building scale.
Following the OVO~\cite{martins2025open} evaluation procedure, we estimate the semantic map classification quality and report localization accuracy on the resulting estimated map in~\cref{tab:replica_metrics}.
We employ the NARADIO~\cite{alama2025rayfronts} image and text encoder for mapping and localization. Ground-truth odometry is perturbed with zero-mean Gaussian noise at 20\% of the per-step motion.
The classification accuracies are reported for all classes (All) and for the most challenging, low-frequency one-third subset (Tail~\cite{martins2025open}). 
The camera poses in OVO~\cite{martins2025open} are estimated using Gaussian-SLAM~\cite{yugay2312gaussian} and ORB-SLAM2~\cite{mur2017orb}, with ATE RMSEs of 0.6\,cm and 1.9\,cm, respectively.

OMCL$_{\textrm{NAR}}$ achieves the best semantic mapping quality due to its high-resolution mapping, while the small improvement over similar RayFronts is likely due to implementation details and design choices.
The largest improvement is observed in the less frequent classes (Tail).
Although the measured semantic quality is high, the subsequent localization on the map is stable but not very accurate (about 35\,cm).
This correlates with the low measurement–map consistency metric estimated for the used dataset–encoder (Replica-NARADIO) pair in~\cref{tab:mapping}.
Replica is smaller but also semantically denser than the Matterport3D dataset, which makes it more challenging to perceive visual features on the map equally well from all viewing directions.
Precision and IoU in~\cref{tab:mapping} correlate with localization accuracy across the Matterport3D and Replica datasets. 

\begin{table}[t]
\centering
\begin{threeparttable}
\caption{ATE on Sequence~00 of the KITTI Dataset.}\label{tab:semantic_kitti_eval}
\label{tab:kitti_ate}
\renewcommand{\arraystretch}{1.1}
\begin{tabular}{lccccc}
\toprule
 & \multicolumn{2}{c}{Translation [m]} & \multicolumn{2}{c}{Rotation [°]} & \multirow{2}{6em}{\centering \textbf{Fine-tuning}\\\textbf{data}}\\
\cmidrule(lr){2-3} \cmidrule(lr){4-5}
 & \textbf{Mean} & \textbf{Std} & \textbf{Mean} & \textbf{Std} & \\ 
\midrule
Stereo DSO~\cite{wang2017stereo} & 7.23 & 5.0 & 1.99 & 0.86 & \textbf{no training} \\
SemLoc~\cite{liang2022semloc} & 1.49 & -- & -- & -- & other urban-driving \\
Zhang \textit{et al.}~\cite{zhang2023cross} & 0.58 & -- & -- & -- & KITTI\\
Pi-Long~\cite{deng2025vggtlongchunkitloop} & 9.88 & 5.39 & 6.24 & 3.61 & \textbf{no fine-tuning} \\ 
\underline{OMCL$_{\textrm{X-Dec}}$}(our) & \underline{0.52} & \underline{0.26} & \underline{1.77} & \underline{0.26} & \textbf{no fine-tuning} \\
\textbf{CMRNext}~\cite{cattaneo2025cmrnext} & \textbf{0.11} & \textbf{0.06} & \textbf{0.25} & \textbf{0.15} & KITTI \textrm{+} other urban\\
\bottomrule
\end{tabular}

\end{threeparttable}
\vspace{-0.3cm}
\end{table}

\textbf{SemanticKITTI.} SemanticKITTI~\cite{BehleyGMQBGS:IJRR21} contains semantically annotated 3D LiDAR from urban driving.
As in SemLoc, we rely on the reconstructed semantic point cloud and evaluate localization on existing maps.
OMCL$_{\textrm{x-dec}}$ employs X-Decoder~\cite{zou2023generalized} as the visual localization model and language encoder, with odometry provided by Stereo DSO.
We use 512 particles, 4096 rays per image, and an Octree Language map resolution of 0.05\,m. Following CMRNext~\cite{cattaneo2025cmrnext}, we report the averaged localization metrics on   Sequence~00 in~\cref{tab:semantic_kitti_eval}.

OMCL$_{\textrm{X-Dec}}$ outperforms other semantic-consistency-based localization methods in accuracy, while requiring no fine-tuning, leading to a significant improvement in the estimated odometry.
CMRNext is the state-of-the-art camera-to-LiDAR map matching approach, but it requires fine-tuning to the data and was trained on the remaining KITTI sequences.
We also compare against the feed-forward Pi-Long approach~\cite{deng2025vggtlongchunkitloop} with loop closure, which, similar to ours, performs monocular localization and does not require data-specific fine-tuning.

\subsection{Prompt-augmented Initialization}
In order to assess the proposed initialization, we randomly distribute particles across the map and report the average number of OMCL steps required to achieve the designated localization accuracy in~\cref{tab:global_loc}.
The single OMCL step corresponds to the weighing of the particles and then resampling.
The prompts are randomly generated from the lists of probable words for each starting location, with each prompt containing three to five words.
We evaluate on the sequences of Matterport3D dataset, and localization can be considered sufficient once it reaches 0.2\,m, since this matches the accuracy of the subsequent pose tracking (\cref{tab:localization}).

\Cref{tab:global_loc} shows that the maximum number of steps needed for localization is 44, with fast convergence once 1\,m accuracy is achieved.
On average, localization requires 29 steps to determine the camera pose, and a minimum of 24 steps is needed to reach 0.1\,m accuracy.

Enhancing global localization with the user prompt according to~\cref{prompt-global-loc} allows us to accelerate localization convergence.
OMCL$^\textrm{0.1\,m / prompt}_\textrm{LSeg}$ used 14 steps on average to achieve an accuracy of 0.1\,m with the fastest case taking 3 steps, accelerating localization convergence by factors of 2.7 and 8, respectively.

\input{images/Number_of_particles}


\subsection{Ablation Studies}\label{experiment:number-of-particles}

We demonstrate the impacts of different parameters and ray sampling strategies in~\cref{fig:Number_of_particles,fig:Numbefr_of_samples}. The plots are constructed by running one trajectory ten times with OMCL$_\textrm{LSeg}$.
According to~\cref{fig:Numbefr_of_samples}, our stratified ray sampling strategy (\cref{method:localization}) outperforms the one with an equal distribution of the rays among the observable categories in both the APE metric and the FPS for the high number of rays.
Moreover, our implementation consumes less GPU memory, allowing it to handle a larger number of rays.
Compared with uniform sampling over the full image, our stratified approach can work even with a small number of sampled rays because it tries to cover all different observable object instances.
It tends to cover the full image without duplicated samples when increasing the number of rays.
The APE variance does not decrease gradually with an increase in rays.
Keeping the number of particles constant, using more rays improves the APE by about $10\%$ compared to $2^8$ rays.
\input{images/Number_of_samples}
\input{tables/global_loc}

~\cref{fig:Number_of_particles} demonstrates that the increase in the number of particles reduces the variance of APE.
It is noticeable that even with a small number of particles, OMCL can still perform pose estimation.
In the beginning, the particles have a higher impact on APE compared to the rays (\cref{fig:Numbefr_of_samples}), but the rays have a smaller initial APE error. 
Increasing the number of particles beyond $2^{11}$ leads to a $10\%$ improvement in APE, but the system may become less stable to the track loss.
Together,~\cref{fig:Numbefr_of_samples,fig:Number_of_particles} help to select parameters with corresponding APE and processed frames per second (FPS) to satisfy the application needs.
We do not include the timing of the used visual model in the presented FPS values because it varies for different models and can be parallelized.


%% file: tables/localization2.tex
\renewcommand{\arraystretch}{1.2}
\setlength{\tabcolsep}{4pt}
\begin{table*}[!t]
\begin{center}
\caption{Absolute Trajectory Error (APE) for Localization on Matterport3D Dataset.}\label{tab:localization}
\vspace{-0.1cm}
\begin{tabular}{llcccccccc}
\hline
                                         &       & \textbf{RMSE} {[}m{]} & \textbf{STD} {[}m{]} & \textbf{Mean} {[}m{]} & \textbf{Median} {[}m{]} & \textbf{min RMSE} {[}m{]} & \textbf{max RMSE} {[}m{]} & \textbf{SSE} {[}$\textrm{m}^2${]} & \textbf{Completed Scens} \\ \hline

\multicolumn{1}{c}{\multirow{6}{*}{\textbf{2D}}} &  \cellcolor{gray!7}\textcolor{black!70}{OMCL$_\textrm{GT}$}  &  \cellcolor{gray!7}\textcolor{black!70}{0.11} &  \cellcolor{gray!7}\textcolor{black!70}{0.07} &  \cellcolor{gray!7}\textcolor{black!70}{0.09} &  \cellcolor{gray!7}\textcolor{black!70}{0.08}              &  \cellcolor{gray!7}\textcolor{black!70}{0.08}        &  \cellcolor{gray!7}\textcolor{black!70}{0.18}       &  \cellcolor{gray!7}\textcolor{black!70}{17.38}          &  \cellcolor{gray!7}\textcolor{black!70}{10/10}               \\
\multicolumn{1}{c}{}                    & \textbf{OMCL$_\textrm{LSeg}$}  &    \textbf{0.15}          &\textbf{0.09}             &\textbf{0.13}              &\textbf{0.12}                &\textbf{0.11}                  &\textbf{0.28}                  &\textbf{33.07}                                & \textbf{10/10}               \\
\multicolumn{1}{c}{}                    & OMCL$^\textrm{auto}_\textrm{LSeg}$  & 0.24             &0.15             &0.18              &0.15                &0.14                  &0.52                  &95.0                                &10/10               \\
\multicolumn{1}{c}{}                    & OMCL$_{\textrm{Open}\backslash\textrm{LSeg}}$ &  0.36            &0.24             &0.27              &0.22                &0.15                  &0.62                  &164.07                                &8/10               \\
\multicolumn{1}{c}{}                    & SeDAR$^{\ast}_\textrm{GT}$~\cite{mendez2018sedar}  &    0.79          &0.46             &0.64              &0.54                &0.24                  &2.33                  &1891.2                                & 6/10               \\
\multicolumn{1}{c}{}                    & SIMP$^{\ast}_\textrm{GT}$~\cite{zimmerman2023constructing}  &    1.39          &1.03             &0.79              &0.94                &0.45                  &2.81                  &2898.15                                &4/10               \\\hline
\multicolumn{1}{l}{\multirow{4}{*}{\textbf{3D}}} & \cellcolor{gray!7}\textcolor{black!70}{OMCL$_\textrm{GT}$}    & \cellcolor{gray!7}\textcolor{black!70}{0.15}           & \cellcolor{gray!7}\textcolor{black!70}{0.09}           & \cellcolor{gray!7}\textcolor{black!70}{0.12}            & \cellcolor{gray!7}\textcolor{black!70}{0.1}              & \cellcolor{gray!7}\textcolor{black!70}{0.1}                & \cellcolor{gray!7}\textcolor{black!70}{0.3}                & \cellcolor{gray!7}\textcolor{black!70}{30.45}           & \cellcolor{gray!7}\textcolor{black!70}{10/10}               \\
\multicolumn{1}{l}{}                    & \textbf{OMCL$_\textrm{LSeg}$}     & \textbf{0.2}              & \textbf{0.1}             &\textbf{0.18}              &\textbf{0.17}                &\textbf{0.17}                  &\textbf{0.41}                &\textbf{59.53}                                & \textbf{10/10}               \\
\multicolumn{1}{l}{}                    & OMCL$_{\textrm{Open}\backslash\textrm{LSeg}}$     &0.42              &0.25             &0.31              &0.25                &0.22                  &0.75                  & 223.32                              & 8/10               \\ 
\multicolumn{1}{l}{} & RTAB-Map~\cite{labbe2019rtab}  &5.23 &2.97 & 4.30              & 3.6                & 0.31               & 8.14                  & 25781.93                              & 10/10               \\ 
\hline 
\end{tabular}
\end{center}
\vspace{-0.5cm}
\end{table*}
\renewcommand{\arraystretch}{1}

%% file: tables/parameters.tex
\begin{table}[t]
  \centering
    \caption{OMCL Parameters.}
    \vspace{-0.1cm}
  \label{tab:parameters}
   {
  \begin{tabular}{@{}lc|lc@{}}
    \toprule
    \textbf{Parameter} & \textbf{Value} & \textbf{Parameter} & \textbf{Value} \\
    \midrule
    Discrepancy threshold $\tau$ & 0.02 & Language feature size $F$ &  512\\
    Number of particles &  1024& Surrounding radius $R$&  2 [m]\\
    Number of rays    &   2048 & Similarity threshold $\rho$ &  0.9\\
    Map resolution &  0.02 [m] & Matching criterion $k$ & 500\\    
    \bottomrule
  \end{tabular}
  \normalsize }
  \vspace{-0.15cm}

\end{table}

%% file: tables/mapping.tex
\renewcommand{\arraystretch}{1.0}
\setlength{\tabcolsep}{4pt}
\begin{table}[t]
\vspace{-0.2cm}
\begin{center}
\caption{Measurements-Map Consistency Evaluation.}\label{tab:mapping}
\vspace{-0.1cm}
\begin{tabular}{lllll}
\toprule
\multicolumn{1}{c}{}      & \multicolumn{1}{c}{\textbf{Accuracy} {[}\%{]}} & \multicolumn{1}{c}{\textbf{Precision} {[}\%{]}} & \multicolumn{1}{c}{\textbf{Recall} {[}\%{]}} & \multicolumn{1}{c}{\textbf{IoU} {[}\%{]}}  \\
\midrule
\multicolumn{5}{c}{Matterport3D~\cite{Matterport3D}} \\
 \addlinespace[0.7ex]
\cdashline{1-5} 
\addlinespace[0.7ex]
\multicolumn{1}{l}{\cellcolor{gray!7}\textcolor{black!70}{OMCL$_\textrm{GT}$}} & \multicolumn{1}{c}{\cellcolor{gray!7}\textcolor{black!70}{86.75}}         & \multicolumn{1}{c}{\cellcolor{gray!7}\textcolor{black!70}{78.96}}        & \multicolumn{1}{c}{\cellcolor{gray!7}\textcolor{black!70}{78.95}}         & \multicolumn{1}{c}{\cellcolor{gray!7}\textcolor{black!70}{67.46}}     \\ 
\multicolumn{1}{l}{\textbf{OMCL$_\textrm{LSeg}$}} & \multicolumn{1}{c}{\textbf{82.74}}             & \multicolumn{1}{c}{\textbf{59.74}}            & \multicolumn{1}{c}{\textbf{54.30}}             & \multicolumn{1}{c}{\textbf{43.35}}     \\ 
\multicolumn{1}{l}{OMCL$^\textrm{auto}_\textrm{LSeg}$} & \multicolumn{1}{c}{80.99}             & \multicolumn{1}{c}{59.13}            & \multicolumn{1}{c}{51.24}             & \multicolumn{1}{c}{41.28}     \\

\multicolumn{1}{l}{OMCL$_{\textrm{Open}\backslash\textrm{LSeg}}$ } & \multicolumn{1}{c}{62.89}             & \multicolumn{1}{c}{38.25}            & \multicolumn{1}{c}{42.91}             & \multicolumn{1}{c}{26.95}   \\ 
 \addlinespace[0.7ex]
\cdashline{1-5} 
\addlinespace[0.7ex]
 \multicolumn{5}{c}{Replica~\cite{straub2019replica}} \\
 \addlinespace[0.7ex]
\cdashline{1-5} 
\addlinespace[0.7ex]
\multicolumn{1}{l}{OMCL$_{\textrm{NAR}}$} & \multicolumn{1}{c}{96.53} & \multicolumn{1}{c}{23.87} & \multicolumn{1}{c}{46.8} & \multicolumn{1}{c}{17.1}   \\
\bottomrule
\end{tabular}
\end{center}
\vspace{-0.2cm}
\end{table}
\renewcommand{\arraystretch}{1}

%% file: images/APE_trajectories.tex
\begin{figure*}[t]
\vspace{0.1cm}  
  \centering
  \includegraphics[width=1.0\linewidth]{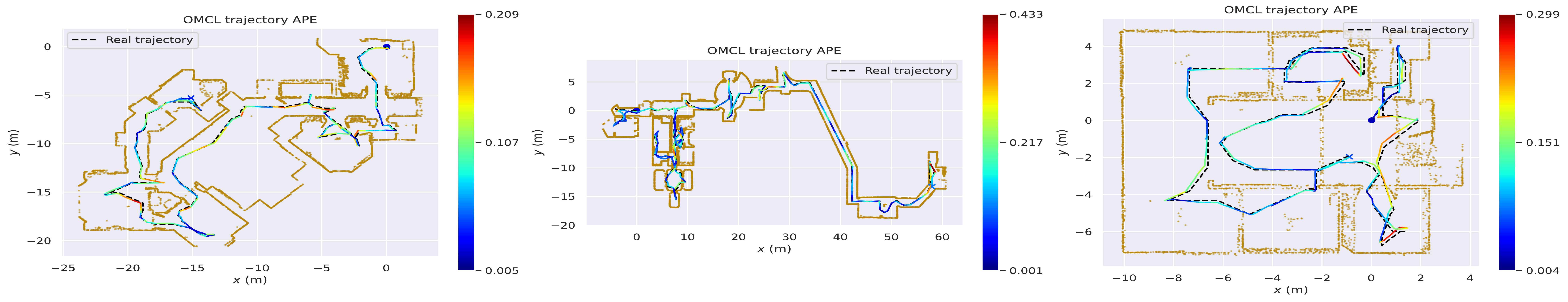}
\vspace{-0.7cm}
   \caption{Example trajectories performed by OMCL on the Matterport3D dataset, with APE indicated by color for each segment. Map projections are shown in brown. The plots demonstrate the performance in scenarios that involve loopy paths, corridors, and long monotonic trajectories.}
   \label{fig:APE_trajectories}
   \vspace{-0.25cm}
\end{figure*}

%% file: images/Number_of_particles.tex
\begin{figure}
\vspace{0.1cm}  
  \centering
  \includegraphics[width=.8\linewidth]{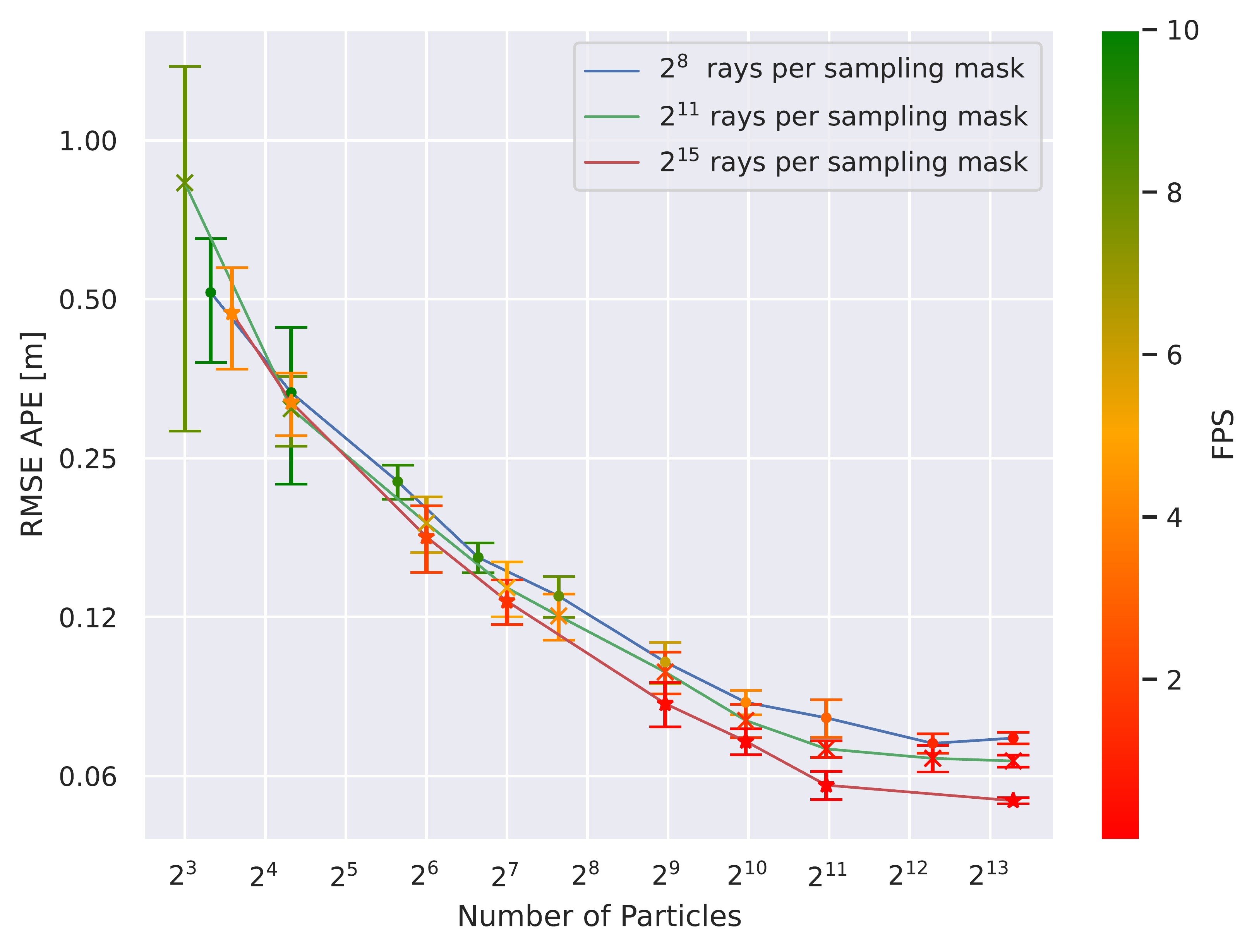}
    \vspace{-0.2cm}
   \caption{Impact of the number of particles on localization accuracy and FPS for varying numbers of sampled rays.
}
   \label{fig:Number_of_particles}
   \vspace{-0.3cm}
\end{figure}

%% file: images/Number_of_samples.tex
\begin{figure}
\vspace{0.1cm}  
  \centering
  \includegraphics[width=.8\linewidth]{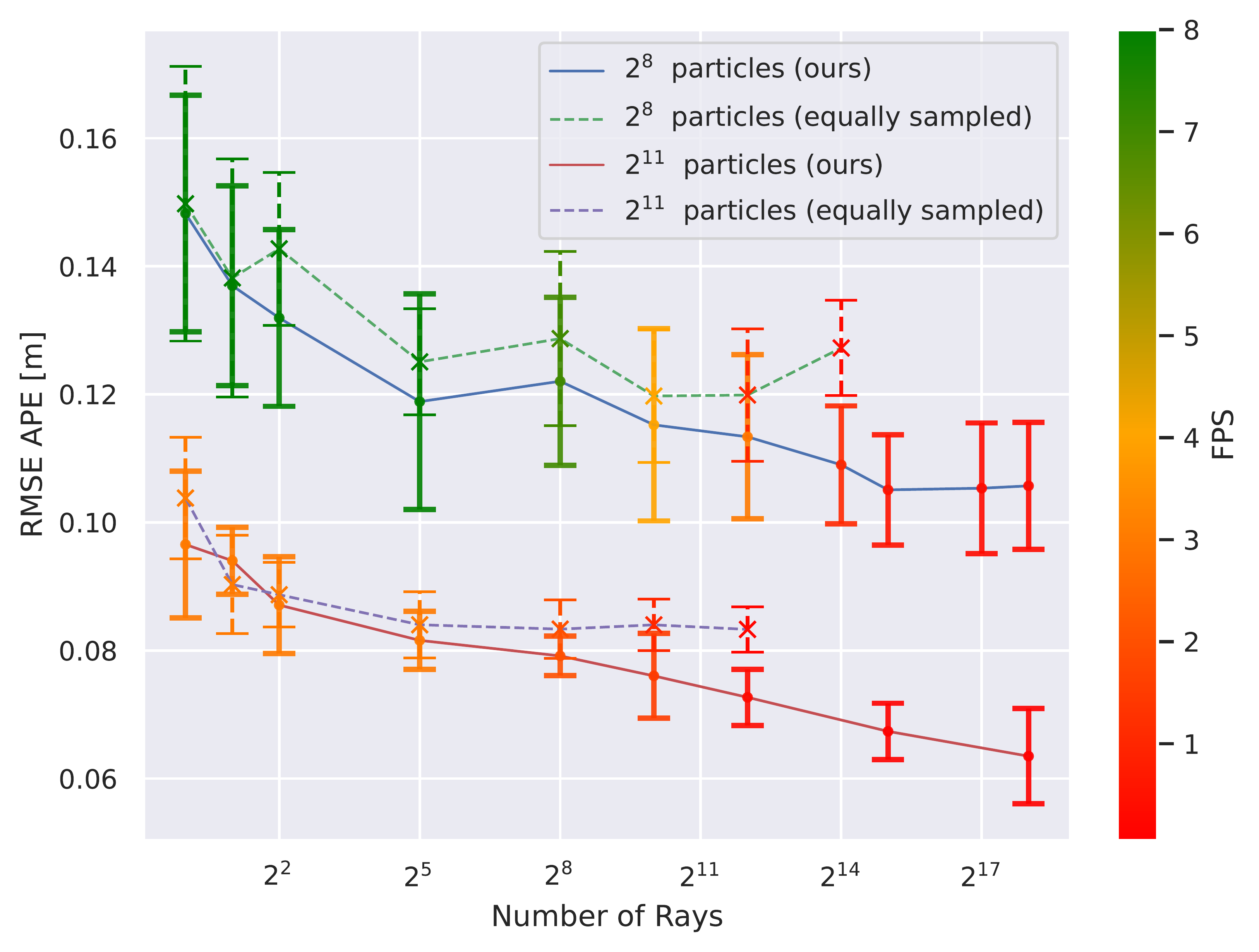}
    \vspace{-0.2cm}
   \caption{
    Comparison between our stratified ray sampling strategy and equal distribution of rays among observable categories, for different particle counts.
}
   \label{fig:Numbefr_of_samples}
\end{figure}

%% file: tables/global_loc.tex
\begin{table}[t]
\vspace{-0.1cm}
\begin{center}
\begin{threeparttable}
\caption{Global Localization Steps Before Convergence.}\label{tab:global_loc}
\vspace{-0.1cm}
\begin{tabular}{ccccc}
 \toprule

{} & {\textbf{mean} [st.]} & {\textbf{std} [st.]} & {\textbf{min} [st.]} & {\textbf{max} [st.]} \\ \midrule
\multicolumn{1}{l}{OMCL$^{\textbf{2\,m}}_\textrm{LSeg}$} & {19} & {8} & {11} & {39} \\ 
 \multicolumn{1}{l}{OMCL$^\textbf{1\,m}_\textrm{LSeg}$} & {22} & {8} & {16} & {41} \\ 
\multicolumn{1}{l}{OMCL$^\textbf{0.5\,m}_\textrm{LSeg}$} & {25} & {8} & {16} & {42} \\ 
\multicolumn{1}{l}{OMCL$^\textbf{0.2\,m}_\textrm{LSeg}$} & {29} & {8} & {17} & {44} \\ 
\multicolumn{1}{l}{OMCL$^\textbf{0.1\,m}_\textrm{LSeg}$} & {38} & {7} & {24} & {45} \\ \bottomrule
\multicolumn{1}{l}{\raisebox{-1mm}{\textbf{OMCL$^\textbf{0.1\,m / prompt}_\textrm{LSeg}$}}} & {\raisebox{-1mm}{\textbf{14}}} & \raisebox{-1mm}{{\textbf{5}}} & \raisebox{-1mm}{{\textbf{3}}} & \raisebox{-1mm}{{\textbf{23}}} \\ 
\end{tabular}
\begin{tablenotes}
\scriptsize
\item[] The steps indicate the number of OMCL iterations needed to reach the specified localization accuracy.
\end{tablenotes}
\end{threeparttable}
\end{center}
\vspace{-0.5cm}
\end{table}
\renewcommand{\arraystretch}{1}

%% file: chapters/5_conclusion.tex
In this paper, we introduced OMCL, a framework for vision-based Monte Carlo localization in open-vocabulary 3D semantic maps.
The maps are created from posed RGB-D images or 3D point clouds and store CLIP features in an octree, enabling cross-modal sensor setups for mapping and localization.
A textual prompt can be used to initialize localization.
OMCL computes the likelihood of localization hypotheses based on the consistency of open-vocabulary features extracted from the current input image and the corresponding map features that are retrieved by ray casting. OMCL is flexible with respect to the choice of the feature extractor and benefits from stratified ray sampling.
Our method generalizes across room-scale indoor, large-scale indoor, and outdoor environments. 
We provide a measurements-map consistency analysis and ablation studies of the proposed framework.

OMCL lacks online map updates and requires map-scaled odometry. Future work on tighter integration with a SLAM backend could potentially address both issues.